\documentclass[runningheads]{llncs}
\usepackage{multirow}
\usepackage[T1]{fontenc}
\usepackage{amsmath}
\usepackage{amssymb}
\usepackage{caption}
\usepackage{subcaption}
\usepackage{booktabs}
\let\oldmarginpar\marginpar
\renewcommand{\marginpar}[1]{\oldmarginpar{\textit{{#1}}}}
\usepackage{graphicx}

\usepackage{hyperref}
\usepackage{color}

\begin{document}
%
\title{Industry Classification Using a Novel Financial Time-Series Case Representation}
\titlerunning{A CBR Approach to Company Sector Classification}
%

\author{Rian Dolphin\inst{1}
\and
 Barry Smyth\inst{1,2} 
 \and
 Ruihai Dong\inst{1,2} 
 }

%
\authorrunning{R. Dolphin et al.}
%
\institute{School of Computer Science, University College Dublin, Dublin, Ireland\\
\email{\href{mailto:rian.dolphin@ucdconnect.ie}{rian.dolphin@ucdconnect.ie}} \and
Insight Centre for Data Analytics, University College Dublin, Dublin, Ireland\\
\email{\{barry.smyth, ruihai.dong\}@ucd.ie}}
\maketitle              
\begin{abstract}
The financial domain has proven to be a fertile source of challenging machine learning problems across a variety of tasks including prediction, clustering, and classification. Researchers can access an abundance of time-series data and even modest performance improvements can be translated into significant additional value. In this work, we consider the use of case-based reasoning for an important task in this domain, by using historical stock returns time-series data for industry sector classification. We discuss why time-series data can present some significant representational challenges for conventional case-based reasoning approaches, and in response, we propose a novel representation based on stock returns embeddings, which can be readily calculated from raw stock returns data. We argue that this representation is well suited to case-based reasoning and evaluate our approach using a large-scale public dataset for the industry sector classification task, demonstrating substantial performance improvements over several baselines using more conventional representations.


\end{abstract}

\section{Introduction}
Case-based reasoning (CBR) approaches have been applied in financial domains, and for a variety of tasks, from the early days of the field; see for example the work of ~\cite{slade1991case} on the use of CBR for financial decision-making. In the years since, there have been many efforts made to apply CBR ideas to a diverse range of financial decision-making and prediction tasks~\cite{kim2004toward,oh2007financial,li2009predicting,wang2016case,chun2020geometric,dolphin2021measuring}. Nevertheless, the use of CBR in such domains is not without its challenges, not the least of which concerns the very nature of many financial datasets and their relationship to the similarity assumption that underpins CBR. The central dogma of CBR is that similar problems have similar solutions, yet financial regulators are always at pains to point out that ``past performance is not a guarantee of future results'' suggesting that this principle may not be so reliable in financial domains, at least when it comes to predicting the future. As society changes and economies ebb and flow, companies that were once the stock market darlings fall out of favour, while new winners seem to emerge, with some regularity, albeit unpredictably and often with little or no warning. Two companies that were similar in the past may no longer be considered similar in the present, while the trajectories of companies with divergent histories might suddenly converge if future circumstances conspire in their favour. All of this greatly complicates the central role of similarity in case retrieval. 

In addition, the feature-based (attribute-value) representations that are commonplace in CBR systems may not provide such a good fit with the type of sequential time-series data (e.g. daily, weekly, and monthly stock prices/returns) that are the norm in financial domains. This is not to say that case-based methods cannot be used with time-series data. Certainly, there is a wealth of literature on representing time-series data for use with case-based methods in a range of application domains from agricultural science~\cite{delaney2022forecasting} to user experience~\cite{lora2017time}. Usually, the approach taken is to use various feature extraction techniques to identify landmark features from the raw time-series data; to effectively transform a raw time-series into a more conventional feature-based representation that can be used with standard similarity metrics. Similar approaches have been applied in the financial domain \cite{kumar2020technical} but, as mentioned above, the stochastic nature of financial markets makes it difficult to effectively isolate useful case representations from market noise, which further complicates the similarity assessment even given a suitable fixed representation.


Thus, in this work, our main technical contribution is to propose and evaluate a novel approach to learning case representations for financial assets (companies/stocks) using raw time-series data made up of historical daily returns. We describe how to transform raw, time-series data into an embedding-style representation of each stock/company; see for example \cite{mikolov2013efficient,nalmpantis2019signal2vec} for examples of embedding representations. We argue that this facilitates the capture of more meaningful patterns and sub-patterns over extended periods of time, while facilitating the type of temporal alignment that is necessary during case comparison and similarity assessment. We argue that this approach is well-suited to the use of case-based and nearest-neighbour approaches in financial domains, because it can be used with a variety of standard similarity metrics, as well as more domain/task specific metrics~\cite{dolphin2021measuring}.  We demonstrate its performance benefits in a comparative evaluation of the industry sector classification task, an important and practical benchmark problem in many financial applications~\cite{phillips2016industry}. 

The remainder of this paper is organised as follows. In the next section, we review the use of case-based reasoning in the financial domain and with time-series data more broadly, highlighting several common tasks and the approaches taken thus far, as well as the important challenges that remain with respect to representation and similarity assessment. Then, in Section \ref{sec:methodology} we discuss the details of our proposed approach, by describing how raw time-series data, such as financial returns, can be transformed into an embedding-based representation that is well suited to case-based approaches. In Section \ref{sec:evaluation} we evaluate our proposed approach by using it to classify companies into their market sectors based on their historical returns data. We present the results of a comprehensive quantitative evaluation, which compares our proposed representation to a variety of baseline and naive approaches. We demonstrate how our embeddings-based representations can offer significant classification improvements, relative to more conventional representations of the raw time-series data. In addition, before concluding with a summary of our findings and a discussion of limitations and opportunities for further work, we present further qualitative evidence in support of the proposed approach, by using our representations to visualise the industry sectors that emerge from a clustering of our cases and some examples of nearest-neighbours in the resulting case-space.

\section{Background}
Case-based reasoning continues to offer many benefits even in a world of big data and deep learning. Its so-called \emph{lazy} approach to problem-solving, which retains the raw cases, offers several benefits when it comes to transparency, interpretability, and explainability \cite{mcsherry2012lazy}. And, the central role of similarity plays -- using similar cases to solve future problems -- helps to lift the computational veil that all too often obscures the computational processes that drive more recent machine learning approaches~\cite{warren2022better}. That being said, the success of CBR is contingent upon the quality of the available cases and the suitability of the case representations and metrics used to evaluate case similarity and drive inference. Case-based reasoning approaches have been particularly effective in domains where cases are plentiful and where feature-based representations are readily available. For example, in loan/credit approval tasks ~\cite{smyth1995comparison}, past decisions provide a plentiful supply of relevant cases, and each case can be represented by salient features such as the value of the requested loan, the debt-load of the applicant, the current earnings of the applicant, the purpose of the loan etc. 


However, in other financial domains the situation is more complex, especially when the available data is sequential/temporal in nature, as is often the case. When it comes to representation, several approaches have been proposed to capture the salient features of financial time-series data, such as asset prices of stock returns. They can be broadly categorised into three groups: (i) traditional feature-based summaries, (ii) raw time-series, and (iii) machine learning-based representations. 

Feature-based representations of financial time-series tend to derive summary features from statistical moments and technical indicators~\cite{gold2015indicators}. For example, time-series data can be represented by extracting key statistical features (e.g. max/min values, mean and standard deviations etc.) over fixed time periods. In addition, the financial domain has the advantage of the availability of several widely-accepted technical indicators, which correspond to common patterns observed in historical trading data such as \emph{on-balance volume}, the \emph{accumulation/distribution line}, the \emph{average directional index}, or the \emph{Aroon indicator}~\cite{gold2015indicators}. These exotic-sounding indicators are among the tools of the trade for technical analysts and day traders and can be readily extracted from financial time series data to provide a valuable source of domain-specific features.

In contrast to feature-based representations, some researchers have explored the use of raw time-series data in CBR applications. Here, instead of computing domain-specific features, the choice of similarity metric accounts for the temporal nature of the data. 
One popular time series similarity technique used in CBR is Dynamic Time Warping (DTW)~\cite{sakoe1978dynamic}, which measures the similarity between two time-series by allowing for temporal shifts in alignment in order to optimise the correspondence between the two time-series. While DTW has been successfully applied in CBR systems across various domains~\cite{delaney2022forecasting}, it is not directly applicable to financial returns data, at least according to the type of tasks that are of interest in this work, because allowing significant temporal shifts in alignment can distort the relationships that exist between two stocks/assets; two stocks having similar returns only constitutes a meaningful relationship if those returns are aligned over a similar same period of time. 
More specifically, in the financial domain, authors in \cite{chun2020geometric} propose a geometrically inspired similarity metric for financial time series, while \cite{dolphin2021measuring} propose a metric combining cumulative returns with an adjusted correlation. 
However, as we show in Section \ref{sec:evaluation}, applying a similarity metric to raw time-series data may not capture all of the relational information needed leading to poorer performance in some tasks.


More recently, so-called \emph{distributed representations}~\cite{mikolov2013efficient} and the use of \emph{embeddings} have become important in the machine learning literature, especially in natural language domains. Embeddings provide a way to translate a high-dimensional representation (such as text) into a low-dimensional representation, which can make it more straightforward to use machine learning techniques when compared with high-dimensional, sparse vectors such as a one-hot encoded vocabulary. Embeddings have been shown to do a good job of capturing some of the latent semantics of the input by locating semantically similar examples close to each other in the embedding space~\cite{mikolov2013efficient}. Indeed they have recently helped to transform many approaches to natural language processing. Similar ideas have been recently explored with financial time series data ~\cite{dolphin2022embeddings,dolphin2023machine} and these have inspired the approach taken in this work. In what follows, we show how to learn case representations, by using the financial returns data of individual companies, and by mapping this high-dimensional raw time-series data into a low-dimensional embedding space. We do this by constructing a similarity-based representation of companies across several time periods and using matrix factorization techniques to compute a low-dimensional representation of these similarity patterns, which can then be used as the basis for our case representation.



\section{An Embeddings-Based Case Representation}\label{sec:methodology}
In this section, we describe the technical details of our approach to transforming raw time-series data into an embeddings-based representation. We will do this using a dataset of stock market \emph{returns} data (see below) for $N=611$ stocks spanning 2000-2018~\cite{dolphin2023machine}. Equivalently, we could use data for other types of financial assets, or more generally a variety of alternative time-series data from other domains. In our evaluation, we use daily, weekly, and monthly returns but the approach described is, in principle, agnostic to granularity. 




\subsection{From Raw Cases to Sub-Cases}
We can consider each complete time-series as a \emph{raw case} so that, for example, $c(a_i)$ corresponds to the full time-series for company/asset $a_i$ as in Equation \ref{eq:rawcase}. Note that in this work the time-series provides so-called \emph{returns} data rather than actual \emph{pricing} data.  The former refers to the relative movement in stock price over a given time period; for example, a return of 0.02 indicates that a price increased by 2\% over a given time period whereas a return of -0.005 indicates that a stock price fell by 0.5\% over a given time period. From this daily returns dataset, we can also aggregate to weekly or monthly returns in a straightforward manner by accumulating returns across longer periods.

\begin{equation}
    \label{eq:rawcase}
    c(a_i) = \{r_{1}^{a_i}, r_{2}^{a_i}, ..., r_{T}^{a_i}\}
\end{equation}

The first step in our approach transforms these raw cases into a set of \emph{sub-cases} such that $c(a_{i}, t, n)$ denotes the sub-sequence of $n$ (the \emph{look-back}) returns ending at time $t$, as shown in Equation \ref{eq:subcase}. For example, later we consider a representation that is based on daily returns with a look-back of five (trading) days (one trading week), which is based on sub-cases with five returns ($n=5$).

\begin{equation}\label{eq:subcase}
    c(a_i,t, n)=\{r_{t-n+1}^{a_i}, r_{t-n+2}^{a_i}, ..., r_{t}^{a_i}\}
\end{equation}

These sub-cases serve as useful and manageable sub-sequences of returns data for the purpose of similarity comparison during the next step.

\subsection{Generating the Count Matrix}\label{sec:count_matrix}
Thus, each company/asset can be transformed into a set of sub-cases and for each asset, look-back duration, and point in time there is a unique sub-case. Next, given a suitable similarity metric, we can produce a $N\times N$ matrix, $\mathcal{S}^{[t,n]}$ of pairwise similarities for any time $t$ and look-back $n$, such that each element is given by $\mathcal{S}_{i,j}^{[t, n]}=sim\big(c(a_i,t, n),c(a_j,t, n)\big)$. Taking stock $a_i$ as an example, we can then use $\mathcal{S}^{[t, n]}$ to identify the stock $a_j$ which is most similar to $a_i$ at time $t$ by finding the column with the maximum value in row $i$ of $\mathcal{S}^{[t, n]}$. 

By repeating this procedure for every $a_i$ and $t$ we can count the number of times that every stock $a_j$ appears as the most similar stock to a given $a_i$, across all time points. The result is a so-called \emph{count matrix} $\mathcal{C}$ such that $\mathcal{C}_{i,j}$ denotes the number of time periods where stock $a_j$ appeared as the most similar stock to $a_i$; see Equation \ref{eqn:count_matrix}. 

\begin{equation}
\label{eqn:count_matrix}
    \mathcal{C}_{i,j}=
    \sum_{\forall t} \delta\left(j
    ~,~
    \arg\max_{\hat{\jmath}\neq i} 
    sim \left(c_{a_i,t},c_{a_{\hat{\jmath}},t}\right)
    \right)
\end{equation}

where $\delta(i,j)$ is the Kronecker delta function as defined in equation \ref{eqn:kronecker}.

\begin{equation}\label{eqn:kronecker}
    \delta(i,j) = \begin{cases}
0 &\text{if } i \neq j,   \\
1 &\text{if } i=j.   \end{cases}
\end{equation}

This approach to computing $\mathcal{C}$ can be viewed as a special case of a more general approach to computing $\mathcal{C}^{[k]}$. Since $\mathcal{C}$ is based on counts that come from the \emph{single} most similar stocks, we can view it as $\mathcal{C}^{[k]}$ where $k=1$. More generally then, $\mathcal{C}^{[k]}_{i,j}$ denotes the number of times where stock $a_j$ appeared among the $k$ most similar stocks to stock $a_i$. In other words, rather than limiting the count matrix to the single most similar stocks we can include a hyper-parameter $k$ to accommodate a more \emph{generous} counting function in order to encode information about a greater number of pairwise similarities. In fact, this is an important practical distinction as our preliminary studies found that representations based on higher values of $k$ performed better during our evaluation. As such for the remainder of this work we will implicitly assume $k=50$, which is the setting used during the evaluation in the next section; we will continue to refer to $\mathcal{C}^{[k]}$ as $\mathcal{C}$, without loss of generality.

In this way, $\mathcal{C}$ tells us about the most similar comparison stocks for a given stock over time. As the time-series data fluctuates to reflect complex, noisy, and unpredictable market changes, different stocks will appear among the \emph{top-k} most similar stocks at different points in time and for different periods of time. We note that every value in $\mathcal{C}$ must be less than or equal to $T$, the number of time points in our raw data, and that cases are not compared to themselves so the diagonal entries in $\mathcal{C}$ are fixed as 0.

\subsection{Generating Embedding Representations}
We can use the count matrix to generate our final case representation by generating an embedding matrix $\mathcal{E}\in\mathbb{R}^{N\times d}$ (randomly initialised) where $d$ is a hyperparameter to determine the the dimensionality of the embedding.
If $\mathcal{E}_i\in\mathbb{R}^{d}$ denotes the $i$\textsuperscript{th} row of $\mathcal{E}$, which represents the embedding for stock $a_i$, then
we can learn the $N\times d$ embedding matrix, $\mathcal{E}$, using matrix factorization techniques by minimising the loss function shown in Equation \ref{eqn:loss} with respect to $\mathcal{E}$. This is related to the problem of learning user and item embedding matrices ($U$ and $V$ respectively) from a rating matrix $R$, in recommender systems, by optimising for $R=UV^T$~\cite{koren2009matrix}, except that here we are producing a case embedding matrix ($\mathcal{E}$) based on $\mathcal{C}=\mathcal{E}\mathcal{E}^T$. However, since we are only optimising a single matrix we must adjust the approach to exclude the diagonal entries of $\mathcal{C}$ from the optimisation.

\begin{equation}\label{eqn:loss}
    \mathcal{L} ~= 
    \sum_{i\in\{1,...,N\}}\sum_{j\neq i} 
    \left(\mathcal{C}_{i,j} - \mathcal{E}_i^T\mathcal{E}_j\right)^2
\end{equation}


To complete the process of learning case embeddings there are a number of routine adjustments that need to be made in order to deal with the type of overfitting and scaling problems that may occur due to the presence of outliers and skew within the distribution of values in the count matrix. First, to prevent the learned embeddings from overfitting to outliers, we define an upper bound for values in $\mathcal{C}$ as the 99.9\textsuperscript{th} percentile of off-diagonal elements in $\mathcal{C}$, clipping any values in $\mathcal{C}$ above this boundary to the boundary; this produces a clipped matrix which we refer to as $\tilde{\mathcal{C}}$. Second, to reduce skew in $\tilde{\mathcal{C}}$ we apply a standard log transformation in Equation \ref{eqn:transformation}. Finally, we apply min-max scaling to the resulting clipped and transformed count matrix, and regularization to the embedding vectors, which gives the final loss function as shown in Equation \ref{eqn:loss_transformed}. In this final loss function. $\mu(\cdot)$ represents the min-max scaling of a matrix over all elements, $f(\cdot)$ represents the log transformation in Equation \ref{eqn:transformation} and $\lambda$ is the regularization rate, which takes a value of 0.1 in our later experiments.

\begin{equation}\label{eqn:transformation}
    f(\mathbf{x}) = \left(\frac{1}{2}\log(1+\mathbf{x})\right)^2
\end{equation}





\begin{equation}\label{eqn:loss_transformed}
    \mathcal{L} ~= 
    \sum_{i\in\{1,...,N\}}\sum_{j\neq i} 
    \left[
    \mu\left(
    f(\tilde{\mathcal{C}}_{i,j})
    \right)
    - 
    \mathcal{E}_i^T\mathcal{E}_j\right]^2
    +
     \lambda \cdot \left( ||\mathcal{E}_i||^2 + ||\mathcal{E}_j||^2 \right)
\end{equation}

\subsection{Discussion}
In summary then, the above procedure transforms a raw, ($N\times T$) times series dataset into a more compact ($N\times d$, where $d<<N$) matrix of embeddings vectors. Each company case corresponds to a single $d$-dimensional embedding vector; that is, a row of $\mathcal{E}$ with its $d$ feature values. This has the advantage of greatly reducing the dimensionality of our cases ($d<<T$) but, in addition, we also hypothesise that the manner in which these embeddings have been produced means that they will capture more useful information than the raw returns data alone, or than more traditional summary features, by surfacing important temporal similarity information about the relationship between stocks. Also, we propose that this information can be usefully applied to good effect in a CBR setting using conventional similarity metrics.

It is worth noting too that the above approach serves as a framework for generating case representations with different levels of granularity, context windows/look-back durations, and different similarity metrics. For example, it may be useful to focus on daily returns over a 5-day look-back period (sub-cases that correspond to single trading weeks) for one task and weekly returns over a 12-week look-back period for a different task. Or it may be useful to consider ways in which the resulting embedding representations can be combined to provide even richer representations. For example, the Orthogonal Procrustes Problem~\cite{schonemann1966generalized} offers a robust solution for aligning embeddings produced by different models. Given two embedding spaces $A$ and $B$, the objective is to find an orthogonal transformation matrix $\Omega$, most closely mapping $A$ to $B$. Mathematically, this can be expressed as the minimization problem $\arg\min_\Omega\|\Omega A-B\|_F$  subject to $\Omega^T$. In principle, such an approach may facilitate combining case representations produced from different sub-cases but we leave this as a matter for future work.

\section{Evaluation}\label{sec:evaluation}
So far, this paper has presented a novel approach for learning embedding-based case representations from financial time-series data, specifically the daily, weekly, and monthly returns data from stocks. We argue that this approach allows us to encode important temporal relationships between financial assets, which are otherwise difficult to capture in more traditional case representations (such as summary, raw feature-based or fixed, attribute-value style representations). In this section, we demonstrate the value of this new approach by evaluating the efficacy of these representations using several qualitative and quantitative techniques. In particular, we provide the results of a comparative evaluation of our embeddings-based representations versus more conventional approaches in a common financial domain classification task.

\subsection{Dataset and Methodology}
In this evaluation, we evaluate the performance of several approaches to industry sector classification using a real-world, publicly available dataset. This is a challenging classification task in its own right, which is instrumental to a multitude of downstream tasks in the financial domain~\cite{phillips2016industry}. As markets and companies evolve historical sector labels may become more or less relevant, which makes the classification task especially challenging, especially when measured with respect to a single target classification label.

It is worth noting that in this evaluation we are primarily focused on using CBR methods for this task and, as such, our main interest is in comparing different representations for use in a CBR system. This is not to say that the resulting systems offer state-of-the-art classification performance. That being said, and as we will discuss later, the use of our embeddings-based representations with CBR does offer a level of classification performance that is competitive with the best available approaches in the literature~\cite{sarmah2022learning,dolphin2022embeddings}, many of which have been optimised for this particular classification task.

\subsubsection{Evaluation Dataset.}
As mentioned previously the dataset used in this work is a publicly available dataset of \emph{returns} data for 611 individual company stocks, spanning the years 2000-2018~\cite{dolphin2022embeddings}. Each stock is associated with a time-series of stock returns (relative changes in price) over daily, weekly, or monthly time periods. The dataset also contains additional (meta) data about each company stock, including industry sector classification data, which will be used in this evaluation. 

\subsubsection{Industry Sector Classification Task.}
For this evaluation we will perform \emph{industry sector classification}, which involves predicting a company's primary industry sector based on their returns time-series data. 
This is a vital task for many types of financial and economic analyses --- identifying peers and competitors, quantifying market share and benchmarking company performance --- none of which would be possible without sector classification schemes~\cite{phillips2016industry}; notably approximately 30\% of publications in the top-three finance journals make use of industry classification schemes~\cite{weiner2005impact}. In this work, our primary focus is to use a case-based reasoning approach to classify stocks, using different representations (see below) to produce different case-base configurations. In each configuration, the problem description of a case corresponds to its returns data (whether using a raw, summary, or embeddings representation) and the solution part of a case corresponds to the stock's sector classification. Then, for a target/query stock $a_q$ we identify its $5$ nearest-neighbours, using a straightforward Euclidean or correlation metric (as given in Table \ref{tab:classification_results}), with simple \emph{majority voting} to identify the predicted industry sector for $a_q$.

\subsubsection{Algorithmic Configurations.} In this evaluation we will test a number of different approaches, each distinguished according to the representation used for cases and the granularity of the returns data (daily, weekly, monthly) used. Arguably the simplest approach is to generate a feature-based representation based on summary features extracted from the raw returns data. These summary features include standard statistical features such as \emph{mean, min, max, volatility, 25th percentile, median, 75th percentile} calculated over the daily, weekly and monthly returns data. These ($\times3$) configurations are referred to as \emph{Summary} in what follows; see the first 3 rows in Table \ref{tab:classification_results}. We also implement two versions using the raw returns data as case representations (\emph{Raw}) one set ($\times3$) uses a Euclidean distance metric (referred to as $E$ in Table \ref{tab:classification_results}) when computing the $k$ nearest-neighbours, and another ($\times3$) uses Pearson's correlation to identify the $k$ nearest-neighbours (referred to as $P$ in Table \ref{}); the latter being a more common similarity metric to use in financial domains. Finally, we test several ($\times18$) varieties of our newly proposed embeddings-based representation (\emph{Embedding}), with varying look-back durations for the daily, weekly, and monthly returns; the final three sections of Table \ref{tab:classification_results}. In particular, we vary the similarity metric used when \emph{computing the count matrix} to look at the effect of using Euclidean distance ($E$) versus Pearson's correlation ($P$) versus the more recent hybrid metric ($H$), which combines Eudclidean distance with a modified correlation component, that was originally proposed for the prediction of returns data in \cite{dolphin2021measuring}. We note that for all of these embedding representations, the dimensionality is chosen as $d=15$ and the standard Euclidean distance metric is used during the subsequent $k$NN classification (with the exception of raw with correlation) to enable a like-for-like comparison with the other baselines.

\subsubsection{Evaluation Metrics.} For each of these 27 different variations, we use a standard 5-fold cross-validation to generate and test the classifications produced. For each variation, we produce a standard \emph{classification report}\footnote{In this work all code is written in Python and uses the standard SciKit Learn implementation of $k$NN, cross-validation, and classification reporting.} which provides \emph{precision} (the ratio of true positives to the sum of true and false positives), \emph{recall} (ratio of true positives to the sum of true positives and false negatives), and F1 (the harmonic mean of precision and recall). The reported values are the weighted average for each class weighted by the number of samples in each class. There are 11 sector classes in the dataset: Basic Industries, Capital Goods, Consumer Durables, Consumer Non-Durables, Consumer Services, Energy, Finance, Health Care, Public Utilities, Technology, Transportation.



\subsection{Results}
The results are presented in Table \ref{tab:classification_results}. Each row corresponds to a specific algorithmic configuration and shows the representation used (\emph{Summary}, \emph{Raw}, and \emph{Embeddings}), the granularity of the returns data (\emph{Daily, Weekly, Monthly}) and the similarity metric used for the final $k$NN classification task (\emph{Euclidean or Correlation}). In addition, for the \emph{Embedding} representations, we also include settings for the relevant look-back periods. Finally, each configuration is associated with an overall weighted precision, recall, and F1 score as mentioned previously.

\begin{table}[!tb]
    \centering
    \caption{Results for the case-based industry sector classification task for each of the 27 variations under consideration.}
    \label{tab:classification_results}
    \begin{tabular}{cccccccc}
    \toprule
        \textbf{Representation} & \hspace{0.07cm}\textbf{$k$NN Metric}\hspace{0.07cm} & \textbf{Granularity} &\hspace{0.07cm}\textbf{Lookback} &\hspace{0.07cm}\textbf{Precision} & \hspace{0.07cm}\textbf{Recall} & \hspace{0.07cm}\textbf{F1} \\
        \midrule
        Summary & $E$ & Daily & --- & 0.11 & 0.15 & 0.12   \\
        Summary & $E$ & Weekly & --- & 0.13 & 0.15 & 0.13   \\
        Summary & $E$ & Monthly & --- & 0.15 & 0.18 & 0.15  \\
        \midrule
        Raw & $E$ & Daily & --- & 0.46 & 0.39 & 0.33  \\
        Raw & $E$ & Weekly & --- & 0.45 & 0.43 & 0.36  \\
        Raw & $E$ & Monthly & --- & 0.39 & 0.40 & 0.33  \\
        \midrule
        Raw & $P$ & Daily & --- & 0.56 & 0.48 & 0.41  \\
        Raw & $P$ & Weekly & --- & 0.50 & 0.49 & 0.42  \\
        Raw & $P$ & Monthly & --- & 0.54 & 0.49 & 0.43  \\
        \midrule
        Embedding + $E$ & $E$ & Daily & 5 & 0.67& 0.62 & 0.63   \\
        Embedding + $E$ & $E$ & Daily & 22 & 0.59 & 0.55 & 0.55  \\
        Embedding + $E$ & $E$ & Weekly & 4 & 0.60 & 0.56 & 0.57  \\
        Embedding + $E$ & $E$ & Weekly & 52 & 0.44 & 0.36 & 0.38  \\
        Embedding + $E$ & $E$ & Monthly & 12 & 0.38 & 0.31 & 0.32  \\
        Embedding + $E$ & $E$ & Monthly & 24 & 0.35 & 0.29 & 0.31  \\
        \midrule
        Embedding + $P$ & $E$ & Daily & 5 & 0.68 & 0.62 & 0.64   \\
        Embedding + $P$ & $E$ & Daily & 22 & 0.67 & 0.64 & 0.64  \\
        Embedding + $P$ & $E$ & Weekly & 4 & 0.66 & 0.60 & 0.61  \\
        Embedding + $P$ & $E$ & Weekly & 52 & 0.46 & 0.39 & 0.41  \\
        Embedding + $P$ & $E$ & Monthly & 12 & 0.51 & 0.46 & 0.47  \\
        Embedding + $P$ & $E$ & Monthly & 24 & 0.50 & 0.42 & 0.44  \\
        \midrule
        Embedding + H & $E$ & Daily & 5 & \textbf{0.69} & \textbf{0.65} & \textbf{0.66}   \\
        Embedding + H & $E$ & Daily & 22 & 0.65 & 0.62 & 0.62  \\
        Embedding + H & $E$ & Weekly & 4 & 0.66 & 0.60 & 0.61  \\
        Embedding + H & $E$ & Weekly & 52 & 0.50 & 0.44 & 0.46  \\
        Embedding + H & $E$ & Monthly & 12 & 0.56 & 0.50 & 0.51  \\
        Embedding + H & $E$ & Monthly & 24 & 0.49 & 0.43 & 0.44  \\
        \bottomrule
    \end{tabular}
\end{table}

A number of performance patterns are evident in these results. The poorest performances are associated with the \emph{Summary} representations (F1$\leq0.15$). This is not surprising given that these representations abstract away a lot of the detail that exists in the returns data. While it may be possible to improve upon these representations, for example by including more domain-specific technical features, they provide a useful naive baseline against which to evaluate the improvements of more sophisticated approaches. The more reasonable \emph{Raw} representations perform considerably better, with F1 values as high as 0.49 found among the variations that use a correlation-based similarity metric, arguable the most popular metric in the financial literature. In general, these \emph{Raw} variations using correlation (\emph{Raw+P}) out-perform the corresponding representations using Euclidean distance (\emph{Raw+E}); the former report with $0.33 \leq F1 \leq 0.36$ compared to $0.41 \leq F1 \leq 0.43$ for the latter. Thus, the \emph{Raw+P} variations serve as a useful baseline against which to evaluate the efficacy of the new embeddings-based representations.

Most of the embeddings-based representations outperform these \emph{Raw+P} baselines, regardless of granularity or look-back duration. And, we note too that shorter look-back periods are associated with better performance than longer look-back periods. As further evidence that correlation-based similarity is more appropriate for financial returns data than Euclidean metrics, we note that the embeddings-based representations that are learned using correlation-based similarity (\emph{Embedding + P} with $0.41 \leq F1 \leq 0.64$) out-perform the corresponding representations that were based on Euclidean distance (\emph{Embedding + E} with $0.31 \leq F1 \leq 0.63$). Furthermore, the novel, hybrid metric introduced by \cite{dolphin2021measuring}, which combines elements of Euclidean distance and correlation, tends to perform as well as, and usually better than, the embeddings-based representations using correlation alone (\emph{Embeddings + H} with $0.44 \leq F1 \leq 0.66$). Indeed, the embeddings produced with this hybrid metric always outperform the $Raw+P$ baseline regardless of granularity and look-back.


\subsection{Discussion}
These results support the hypothesis that the proposed embeddings-based representations are capable of capturing more useful information from the time-series returns data than more conventional representations. The best embeddings-based representation is associated with an F1 score of 0.66 compared to just 0.49 for the best baseline representation. Moreover, the proposed representation is well-suited for use in a case-based reasoning setting which offers further advantages when it comes to transparency, interpretability, and explainability.

\begin{table}[t]
    \centering
    \caption{Examples of top-3 nearest neighbours for given query stocks}
    \label{table:example}
    \begin{tabular}{cccc}
        \textbf{\begin{tabular}[c]{@{}c@{}}Query Stock\\ Sector - Industry\end{tabular}} & \textbf{3 Nearest Neighbours - Sector - Industry} & \textbf{Similarity}  \\ \midrule
        \begin{tabular}[c]{@{}c@{}}JP Morgan Chase\\ Finance \\Major Bank\end{tabular}          & \begin{tabular}[c]{@{}c@{}}Bank of America Corp - Finance - Major Bank \\ State Street Corp - Finance - Major Bank  \\ Wells Fargo \& Company - Finance - Major Bank\end{tabular}  & \begin{tabular}[c]{@{}c@{}c@{}}0.98\\0.98\\0.97\end{tabular} \\\hline
        
        \begin{tabular}[c]{@{}c@{}}Microsoft\\ Technology \\ Software\end{tabular}          & \begin{tabular}[c]{@{}c@{}} IBM - Technology - Computer Manufacturing  \\ HP - Technology - Computer Manufacturing \\ Adobe - Technology - Software\end{tabular} & \begin{tabular}[c]{@{}c@{}c@{}} 0.95 \\ 0.93 \\ 0.92\end{tabular} \\ \hline
        
        \begin{tabular}[c]{@{}c@{}}Walmart\\ Consumer Services \\  Department Store\end{tabular}          & \begin{tabular}[c]{@{}c@{}}Costco - Consumer Services - Dept Store \\ Kroger -  Consumer Services - Food Chains \\ McDonalds - Consumer Servies - Food Chains  \end{tabular} & \begin{tabular}[c]{@{}c@{}c@{}}0.89\\0.82\\0.78\end{tabular}  \\ \hline
    \end{tabular}
\end{table}

By way of further explanation, Table \ref{table:example} shows some examples of the nearest neighbours that are identified for 3 different query companies (JP Morgan Chase, Microsoft and Walmart) using an embeddings-based representation. For each query company, we summarise the top-3 nearest neighbours, the sector class (e.g. Finance), a finer-grained industry label (e.g. Major Bank), and their corresponding similarities to the query stock. The results  align with our intuitions and in each case, the nearest neighbours match the query's industry sector (Finance, Technology, and Consumer Services, respectively).

\begin{figure}[tb]
    \centering
    \includegraphics[width=0.8\linewidth]{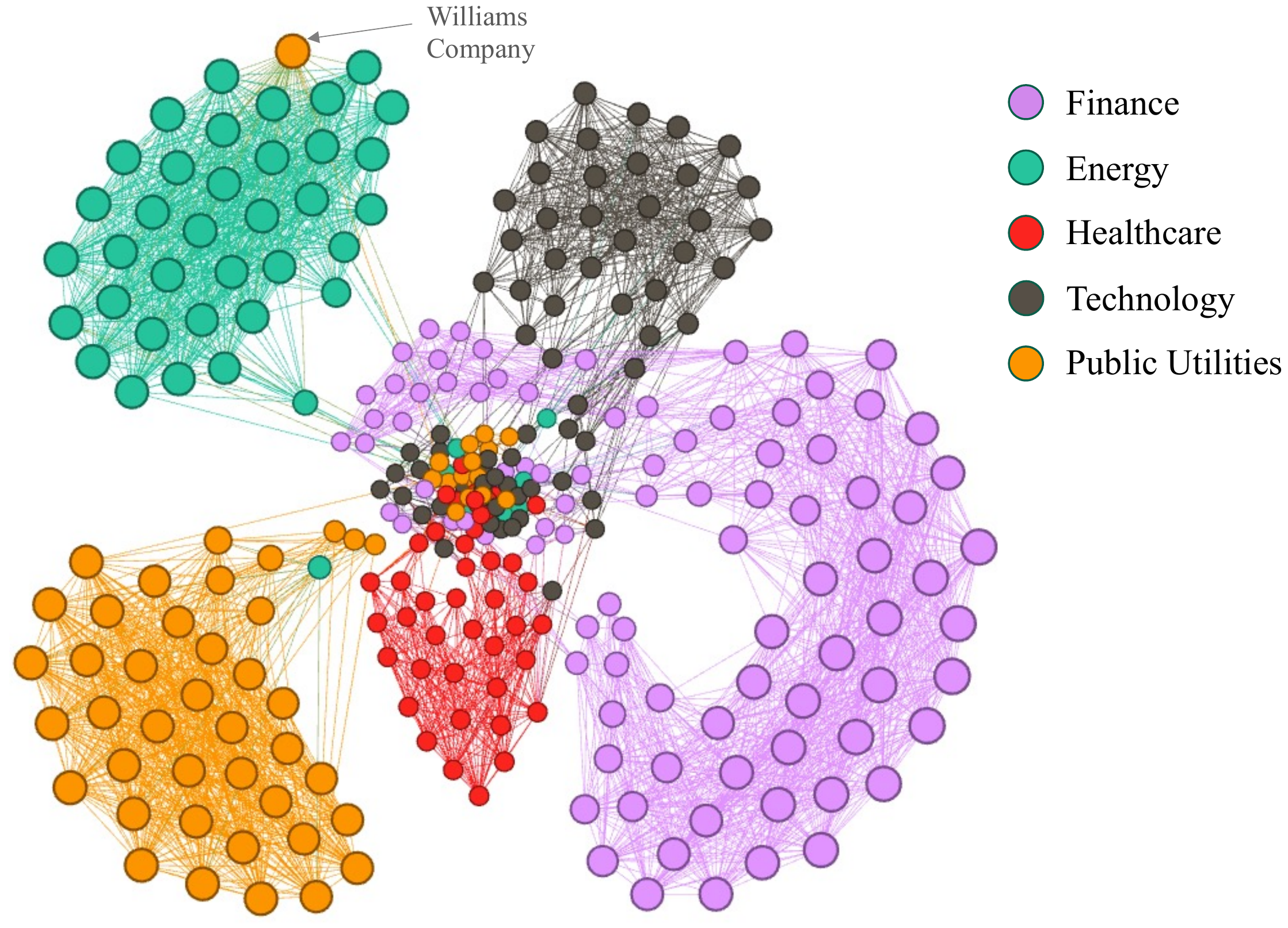}
    \caption{Visualization of embedding clustering. A subset of sectors is used for visual clarity.}
    \label{fig:gephi}
\end{figure}

As another example, Figure \ref{fig:gephi} shows a 2D visualisation of the clusters of companies that emerge when using the embeddings-based representations. Each node corresponds to an individual stock and an edge is created between two stocks if their similarity exceeds some minimum threshold (0.75 in this example). Then, a force-directed graph drawing algorithm~\cite{jacomy2014forceatlas2,SmythMM00} is used to position the nodes in such as way as to optimise their placement in the resulting similarity space. The nodes have been colour-coded based on their ground-truth industry sectors and we can see clearly how nodes of from the same industry sector tend to be clustered together, indicating that the embeddings-based representation is doing a good job of capturing this relationship; as an aside it is worth noting that the embedding representations also exhibited clear clustering using visualisation approaches such as PCA
and t-SNE.



We also observe some interesting patterns in the graph that are not immediately obvious from the sector labels. For example, a node from the Public Utilities sector (indicated in orange) appears as an outlier in the Energy cluster (green). This node, highlighted in Figure \ref{fig:gephi}, is an energy supply company called Williams Company, whose primary business is natural gas processing and transportation. Thus, although it has Public Utilities classification in our dataset, the case representation facilitates recognising its similarity with the Energy business and it is positioned accordingly. 

Visualisations such as this are powerful tools for technical analysts to better understand the evolving structure of modern markets, but to be useful they must rely on representations that are capable of reasonably accurately capturing meaningful relationships between different stocks and companies. The case representation proposed here should help to improve the utility of such tools because it does a better job at recognising the relationships that exist between companies but that may be obscured by the raw time-series data and not captured by traditional subjective industry classification schemes.

\section{Conclusion}
Using case-based reasoning with time-series data presents a number of challenges, not the least of which is how to generate case representations that are capable of capturing the complex temporal behaviour of the underlying data. Time-series data are becoming more and more common in the modern world with the increasing ability to capture and store large amounts of real-time, real-world data. This is especially true in the financial domain and in this work, we have described the development of a novel representation of financial time-series data that is well suited to case-based reasoning. We have  demonstrated the effectiveness of this representational approach on the important task of industry sector classification, in comparison to a number of naive and more reasonable alternative representations. The results indicate that the proposed approach offers some significant benefits compared to these alternatives.

There are several opportunities for additional work arising from this initial study. For example, no comprehensive hyper-parameter tuning has been carried out for the proposed representations and it is likely that by varying key parameters, such as the embedding dimensionality ($d$), $k$, and $\lambda$, that further improvements will be likely; the fact that significant improvements were obtained for the ``default'' settings used here speaks to this. And, although the focus of this work has been on the use of the proposed representation in a case-based reasoning context, the representation should be equally applicable as a training data representation for other machine learning models. In fact, preliminary results, not provided here for reasons of space and clarity, suggest performance that is at least as good as, and often better than, the current state-of-the-art in purpose-built, fine-tuned industry sector classification models.

Within the financial domain, there are many other tasks that can be explored as targets for this type of representation. For example, risk management and portfolio optimisation~\cite{dolphin2022embeddings} are obvious candidates in this regard. Moreover, given the proliferation of time-series data across many domains (clinical health \cite{delaney2021instance}, exercise and fitness \cite{feely2020using}, gaming~\cite{lora2017time} etc.) it will be interesting to assess whether this type of representation can add value across different task types.

%
%
%
\bibliographystyle{splncs04}
\bibliography{main}

\end{document}